\documentclass[lettersize,journal]{IEEEtran}
\usepackage{amsmath,amsfonts}
\usepackage{algorithmicx}
\usepackage{algpseudocode}
\usepackage{algorithm}
\usepackage{array}
\usepackage[caption=false,font=normalsize,labelfont=sf,textfont=sf]{subfig}
\usepackage{textcomp}
\usepackage{stfloats}
\usepackage{url}
\usepackage{verbatim}
\usepackage{graphicx}
\usepackage{graphbox}
\usepackage{cite}
\usepackage{xcolor}
\usepackage{mwe}
\usepackage{amsfonts}
\usepackage{boldline}
\usepackage{xtab}
\usepackage{multicol}
\usepackage{tabularx}
\usepackage{multirow}
\usepackage{courier}

\definecolor{promptcolor}{HTML}{007ACC}
\definecolor{answercolor}{HTML}{D85F00}
\newenvironment{prompt}{
    \begingroup
    \color{promptcolor}
    \small
    \fontfamily{pcr}\selectfont
    \textbf{Prompt:}\\
}{
    \endgroup
}

\newenvironment{answer}{
    \begingroup
    \color{answercolor}
    \small
    \fontfamily{pcr}\selectfont
    \textbf{Response:}\\
}{
    \endgroup
}

\newcommand{\explanation}[1]{\textcolor{black}{#1}}

\begin{document}

\title{SAGE: Smart Home Agent with Grounded Execution}

\author{Dmitriy Rivkin, Francois Hogan, Amal Feriani, Abhisek Konar, Adam Sigal, Xue Liu, Gregory Dudek}

\maketitle
\begin{abstract}
The common sense reasoning abilities and vast general knowledge of Large Language Models (LLMs) make them a natural fit for interpreting user requests in a Smart Home assistant context. LLMs, however, lack specific knowledge about the user and their home limit their potential impact. SAGE (Smart Home Agent with Grounded Execution), overcomes these and other limitations by using a scheme in which a user request triggers an LLM-controlled sequence of discrete actions. These actions can be used to retrieve information, interact with the user, or manipulate device states. SAGE controls this process through a dynamically constructed tree of LLM prompts, which help it decide which action to take next, whether an action was successful, and when to terminate the process. The SAGE action set augments an LLM's capabilities to support some of the most critical requirements for a Smart Home assistant. These include: flexible and scalable user preference management (\textit``is my team playing tonight?"), access to any smart device's full functionality without device-specific code via API reading (\textit{``turn down the screen brightness on my dryer"}), persistent device state monitoring (\textit{``remind me to throw out the milk when I open the fridge"}), natural device references using only a photo of the room (\textit{``turn on the light on the dresser"}), and more. We introduce a benchmark of 50 new and challenging smart home tasks where SAGE achieves a 75\% success rate, significantly outperforming existing \textit{LLM-enabled} baselines' (30\% success rate).

\end{abstract}

\begin{IEEEkeywords}
Autonomous LLM Agents, Smart Home, IoT,  Generative AI, Embodied AI, Personalized AI, AI Assistant.
\end{IEEEkeywords}

\section{Introduction}

This article presents an approach for unlocking the full potential of LLMs for smart home assistants via LLM augmentations within a sequential action framework. SAGE (Smart Home Agent with Grounded Execution) guides this sequential process via a dynamically generated tree of LLM prompts which integrate relevant information critical to a correct interpretation and execution of a user's request. This is referred to as an ``autonomous LLM agent" \cite{LLAsSurvey} strategy because it allows for a higher degree of LLM autonomy than more rigid earlier frameworks, such as \cite{king2023sasha} and \cite{zapier}, where LLMs are used to answer one or more questions within a pre-defined decision making pipeline, but are not able to guide the \textit{control flow} of the process. Actions in the available action set are referred to as ``tools."

Tools have both inputs (arguments) and outputs. An LLM is prompted to decide which tool (and arguments) to use. Internally, a tool may query a database, control a device, interact with the user, or any number of other behaviors, including building and executing other LLM prompts. When the tool finishes execution, the outputs of the tool are used to construct the subsequent prompt. SAGE effects LLM augmentation through the construction of tool selection prompts, tool interfaces, tool implementations, and tool output formats. With these augmentations, SAGE is able to overcome inherent shortcomings of LLMs including but not limited to a lack of specific knowledge about the user, their home, and their devices. The overall approach is described in more detail in Section \ref{sec:system_overview}. Section \ref{sec:tools} introduces the most important tools, and describes their function and implementation.

\begin{figure}[th!]
    \centering
    \includegraphics[width=.49\textwidth]{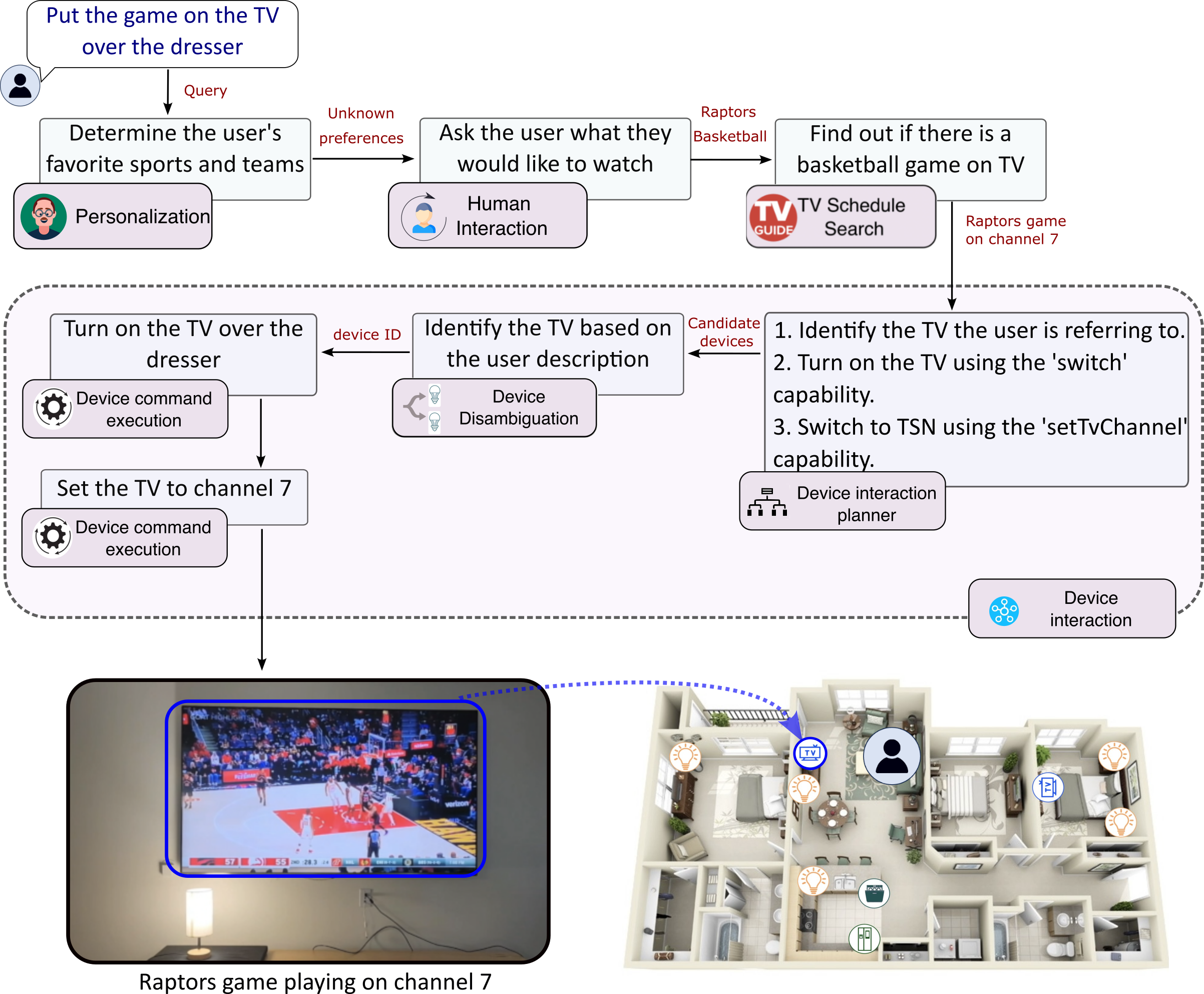}
    \caption{\textbf{Interactive demo where the  user asks SAGE: \textit{``Put on the game by the dresser"}}. The figure illustrates the sequence of tools used by SAGE to complete the task. Note that each control flow decision was made by an LLM, not by hand-coded logic. The control flow is described step by step in Section \ref{sec:execution_example}. This demo was executed using real SmartThings-enabled devices.}
    \label{fig:interactive_demo}
\end{figure}

The upshot of the SAGE design is a decision making flow controlled primarily by an LLM but not limited by its inherent limitations, enabling an unprecedented level of flexibility and naturality in users' interaction with their smart homes. An example of such a decision process is illustrated in Figure \ref{fig:interactive_demo}. This illustration may appear mundane first glance, but the key point is that each step was decided by an LLM, with no manually defined decision logic whatsoever. Because no manual decision logic exists, users are not limited to the use cases envisioned by designers. As such, SAGE represents a significant paradigm shift from legacy solutions (such as Bixby, Alexa, Google Assistant, etc.) and even the next generation of LLM-enhanced versions of such products (e.g. \cite{king2023sasha}).

With SAGE, we aim to enhance user experience by moving beyond the simplistic, highly structured, explicit language common in today's smart required by today's smart home assistant, therefore we designed an evaluation suite to test its ability to do so. This benchmark is composed of 50 challenging tasks, along with test harnesses and validation logic to allow the tests to run without hardware (or human evaluators) in the loop. Details are included in Section \ref{sec:evaluation}. We evaluate SAGE with a number of LLMs and compare it to two LLM-enabled baselines in Section \ref{sec:results}, and demonstrate a performance improvement of over 2x

The main contributions of this paper are:

\begin{enumerate}
    \item \textbf{Personalization tool:} A module for answering questions about user preference based on user interaction history.
    \item \textbf{Device interaction tool:} A module for interacting with any smart device by prompting an LLM with its API documentation.
    \item \textbf{Device disambiguation tool:} A mechanism for allowing users to refer to their devices in a natural and non-rigid way, as they normally would to another human, based on a single photo of that device in the home.
    \item \textbf{Persistent command handling via code-writing:} a scheme for allowing the LLM to persistently monitor device states by exploiting LLM code-writing abilities. 
    \item \textbf{SAGE:} The integration of these tools and others into a coherent system with a single entry point, thereby overcoming key inherent limitations of LLMs within the smart home context.
    \item \textbf{Performance benchmark:} A benchmark of 50 tasks to validate SAGE's performance.
\end{enumerate}

\section{System Overview}
\label{sec:system_overview}

The structure of SAGE's sequential decision-making process is described in Algorithm \ref{alg:hierarchical_agent}. Some tools are themselves implemented as sequential decision making processes which select other tools. As previously indicated, such processes are often referred to as ``agents," e.g. in \cite{LLAsSurvey}. Therefore, we refer to tools which are implemented through such a sequence as ``agent-tools", with a single top-level agent-tool providing the entry point for user interaction.

\begin{algorithm}
\caption{SAGE Decision Process}
\begin{algorithmic}[1]
\Require Set of all tools $T = \{t_1, t_2, \ldots, t_n\}$.
\Require Subset of agent-tools $M = \{m_0, m_1, \ldots, m_p\} \subseteq T$ with $m_0$ as the fixed entry point agent-tool.
\Require Set of decision functions $D = \{d_0, d_1, \ldots, d_p\}$, where $d_i$ corresponds to $m_i$.
\Require Subsets $S_i \subseteq T$, the tools available to agent-tool $m_i$.
\Require Universal set of possible arguments $A$.
\Require Special terminate action $\tau$ indicating a agent-tool should terminate its process and return its output.

\State Initialize decision history $H_0$ for the entry point agent-tool's decision function $d_0$.

\Loop \Comment{Infinite loop awaiting user input}
    \State Await and receive user input $\textit{userInput}\in A$
    \State $output \gets \Call{CallAgentTool}{m_0, H_0, \textit{userInput}}$
    \State Respond to user with $output$.
\EndLoop

\Statex

\Function{CallAgentTool}{$m_i$, $H_i$, $a \in A$}
    \Loop
        \State $(action, output) \gets d_i(H_i, input)$ \If{$action = \tau$}
            \State $H_i \gets empty list$
            \State \Return $output$ \Else
            \State Parse action as $(s, a^*)$ where $s \in S_i$, $a^* \in A$
            \If{$s \in M$} 
                \State Identify the index $j$ such that $s = m_j$
                \If{not $H_j$ initialized}
                    \State Initialize $H_j$ as an empty list
                \EndIf
                \State $o \gets \Call{CallAgentTool}{m_j, H_j, a^*}$
            \Else
                \State $o \gets \Call{s}{a^*}$
            \EndIf
            
        \State $H_i \gets H_i + [(s, a, o)]$
        \EndIf
    \EndLoop
\EndFunction

\end{algorithmic}
\label{alg:hierarchical_agent}
\end{algorithm}

The functions $d_i$ are implemented using an LLM. The LLM prompt \( \mathcal{P}_i \) for the decision function \( d_i \) is constructed as follows:

\begin{multline*}
\mathcal{P}_i = \text{Concatenate}(\texttt{TaskInfo}_i, \texttt{ToolInstructions}(S_i),\\
\texttt{FormatInfo}, a,\texttt{HistoryInfo}(H_i))
\end{multline*}

where:
\begin{itemize}
\item $\texttt{TaskInfo}_i$ provides contextual information about the task to be performed by the meta-tool.
\item  $\texttt{ToolInstructions}(S_i)$ includes instructions on how to use each tool in the subset $S_i$ of tools available to meta-tool $m_i$.
\item $\texttt{FormatInfo}$ specifies instructions on how the output should be formatted to facilitate parsing.
\item $a$ is the input received by the meta-tool.
\item $\texttt{HistoryInfo}(H_i)$ compiles the decision history of the meta-tool into a format that can be understood by the LLM.
\end{itemize}

Given the prompt \( \mathcal{P}_i \), the LLM is then sampled:

\[
(response, probability) \gets \text{SampleLLM}(\mathcal{P}_i)
\]

Here, \( \text{SampleLLM} \) represents the process of querying the language model with the generated prompt \( \mathcal{P}_i \). The language model then returns a response along with an associated probability indicating the confidence of the model in its generated response. The output generated by the LLM needs to be parsed into a format suitable for the algorithm:

\[
(action, output) \gets \text{ParseLLMOutput}(response)
\]

\( \text{ParseLLMOutput} \) is a function that interprets the response from the LLM into a defined \( action \) (either \( \tau \) or a tuple \( (s, a) \) where \( s \) is a tool and \( a \) is an argument) and associated \( output \) data that is meaningful within the context of the algorithm. If the LLM fails to follow the instructions in $\texttt{FormatInfo}$ this is not possible and an error is thrown.

Combining these steps the decision function \( d_i \) using an LLM can be encapsulated as:

\[
d_i(H_i, input) = \text{ParseLLMOutput}(\text{SampleLLM}(\mathcal{P}_i))
\]

Within this framework, planning (the process of breaking down a high-level goal into sub-steps) is implemented using a technique called ``chain-of-thought" \cite{wei2022cot}. In this approach, $\texttt{ToolInstructions}$ and $\texttt{FormatInfo}$ encourage the LLM to output a plan before proceeding to specify exactly how to execute the steps of the plan. An highly simplified example is provided in Figure \ref{fig:prompt_example}. Observe that the multi-step plan output by the decision function ($d_i$) the first time it is called is persisted via the history ($H_i$) so the agent-tool can continue to execute the same plan through multiple LLM calls. It may also alter the plan in response to unexpected results, facilitating failure recovery.
\begin{figure}

\begin{prompt}
\explanation{[TaskInfo]}\\ You are an AI that helps the user turn their lights on and off. Think step-by-step about what to do before you execute.\\\\
\explanation{[ToolInstructions]}\\Tools:
\begin{itemize}
    \item light ID tool (input: light common name)
    \item turn on tool (input: light uuid)
    \item turn off tool (input: light uuid)
\end{itemize}
\explanation{[FormatInfo]}\\Use the following output format\\
Thought: the steps you need to execute to handle the request\\
Action: the next tool to use OR $\tau$\\
Action input: the argument to the tool OR final response (if action is $\tau$)\\\\
\explanation{[$a$]}\\
User input: Turn on the fancy light \\\\
\explanation{[HistoryInfo]}\\
\explanation{Previous LLM response}\\
Thought: First I need to use light ID tool to find the uuid of the fancy light, then I need to turn it on using the turn on tool.\\
Action: light ID tool\\
Action input: fancy light\\
\explanation{Previous tool output}\\
Observation: the UUID of "fancy light" is 12e4df...bc4
\end{prompt}
\begin{answer}
Thought: I now know the UUID of fancy light. Now I need to use light on tool to turn it on.\\
Action: turn on tool\\
Action input: 12e4df...bc4\\
\end{answer}

\caption{Simplified LLM prompt construction and response example. Blue text indicates prompt, orange text indicates response, and black text is explanation of the section of the prompt below. In this example, the decision function ($d_i$) is being called for the second time, so the history ($H_i$) is populated with the results of the first call.}
\label{fig:prompt_example}
\end{figure}

In this work, FormatInfo and ParseLLMOutput are taken from ReAct \cite{yao2022react}. The HistoryInfo function is a concatenation operation. The set of possible arguments, $A$, is the set of all strings. TaskInfo, ToolInstructions, the implementations of all tools in set $T$, and the assignment of sub-tools to meta-tools $S_i$ comprise the method referred to as SAGE. Implementations for several critical tools are described in Section \ref{sec:tools}.

\section{Tools}
\label{sec:tools}
\begin{table*}[ht]
\centering
\caption{SAGE tool hierarchy. Each row corresponds to a tool in SAGE and indicates its functionality category, name, whether it is a tool-agent, and which sub-tools (if any) it has access to.}
\label{table:tools}
\begin{tabularx}{\textwidth}{|lllX|}
\hline
Category & Name & Agent-Tool & Sub-Tools \\
 \hline
entry point & SAGE & yes & personalization, device interaction, condition code writing, condition polling, email and calendar, weather, TV schedule search \\
\hline
\multirow{2}{*}{personalization} & personalization  & no & N/A \\

 & human interaction & no & N/A\\
\hline

\multirow{6}{*}{device interaction} & device interaction & yes & device interaction planner, API documentation retrieval, device attribute retrieval, device command execution, device disambiguation \\
 
& device interaction planner & no & N/A  \\
 
& API documentation retrieval & no & N/A \\
 
& device attribute retrieval & no & N/A \\
 
& device command execution & no & N/A \\
 
& device disambiguation & no & N/A \\

\hline

\multirow{3}{*}{monitoring} & condition code writing  & yes & device interaction planner, API documentation retrieval, device disambiguation, code execution \\

& code execution  & no & N/A\\

& condition polling  & no & N/A\\

\hline

\multirow{11}{*}{external interaction} & email and calendar & yes & get contacts, create calendar event, list calendar events, create email draft, send email message, search email, get email message, get email thread \\

& get contacts & no & N/A \\

& create calendar event & no & N/A \\

& list calendar events & no & N/A \\

& create email draft & no & N/A \\

& send email message & no & N/A \\

& search email & no & N/A \\

& get email message & no & N/A \\

& get email thread & no & N/A \\

& weather & no & N/A \\

& TV schedule search & no & N/A\\

\hline
\end{tabularx}
\end{table*} In this section, we introduce a collection of tools developed for SAGE (see Table~\ref{table:tools} for a comprehensive list). In addition to listing all of the tools, this table indicates which tools are agent-tools, as well as the sub-tools for those agent-tools, thereby providing an overview of the entire tool hierarchy. In order to help organize these tools, we group them into $4$ functionality categories: personalization (accounting for user preferences), device interaction (interacting with smart devices), monitoring (continuous monitoring of device states in order to handle persistent commands), and external interaction (interaction with APIs external to the smart home, e.g. email, weather).

The rest of this section provides implementation details for tools in the personalization, device interaction, and monitoring categories. Details on external interaction tools are omitted for brevity, as these types of tools have been well explored in previous work (e.g. in Langchain \cite{langchain}).

\subsection{Personalization}
\label{sec:personalization}
\begin{figure}[t]
    \centering
    \includegraphics[width=0.48\textwidth]{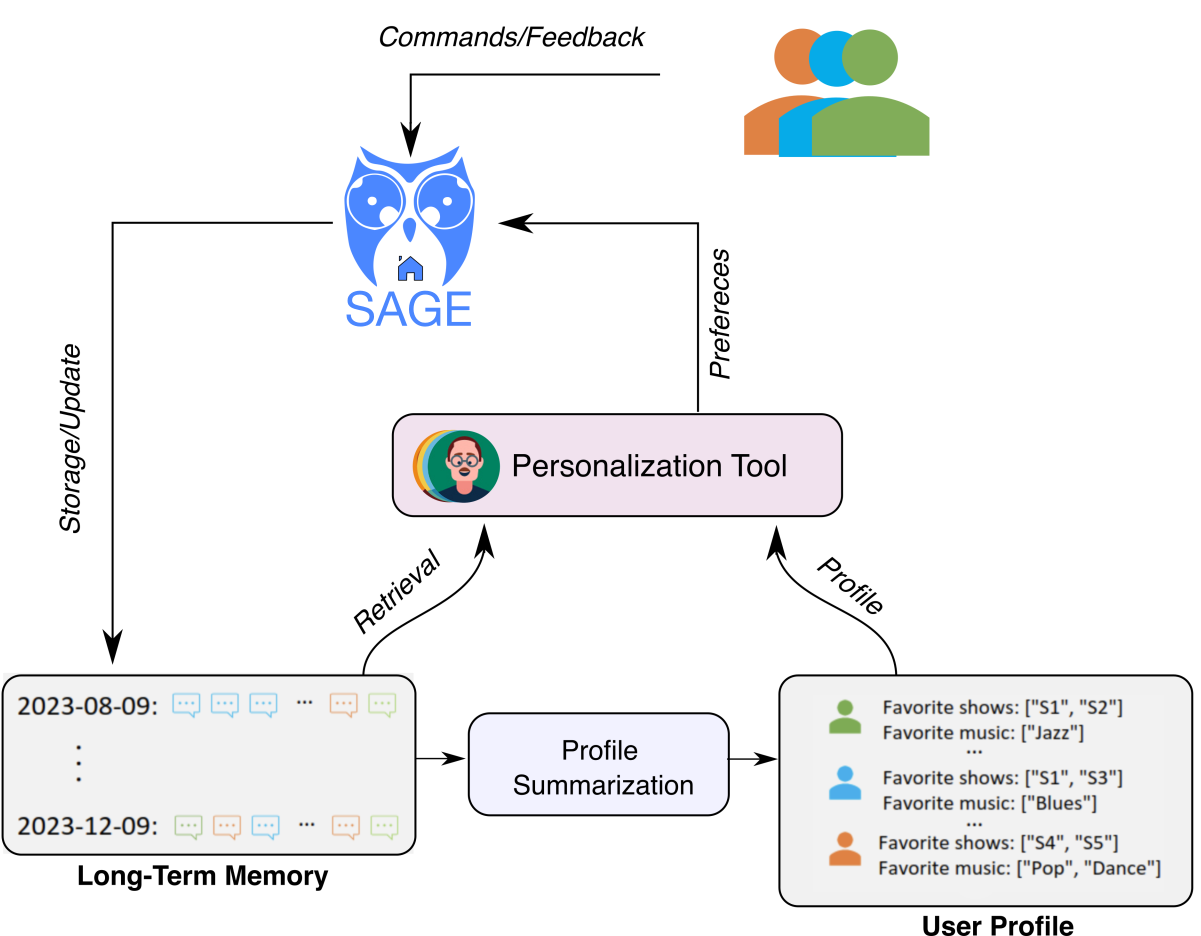}
    \caption{An overview of the personalization tool. Long-term memory remembers and retrieves all past user utterances. User profile contains more general trends about the user's preferences.}
    \label{fig:personalization_module}
\end{figure}

This section introduces tools with personalization-related functionality: the personalization and human interaction tools. The personalization tool, illustrated in Figure~\ref{fig:personalization_module}, is composed of two main sub-components: (i) the long-term memory that stores the history of all past user interactions, and (ii) the user profiler that constructs a hierarchical understanding of user preferences. We first describe these two components, then describe how they are integrated. Finally, we conclude with a brief description of the human interaction tool.

\subsubsection{Long-Term Memory}
Long-term memory \cite{LLAsSurvey}, shown on the bottom-left of Figure~\ref{fig:personalization_module}, stores information about the user's past interactions and behavior. The memory records a history of user commands commands and feedback. Similar to existing information retrieval techniques for LLM augmentation \cite{zhu2023IRsurvey}, the long-term memory is used to retrieve memories relevant to the user query in order to augment the in-context information available to the personalization tool. Each entry in the memory is encoded using a dense retrieval embedding model \cite{karpukhin2020DPR}. The vector representations are then indexed and stored in a vector database. In this work, we use the MiniLM embedding model \cite{miniLM}.

\subsubsection{User Profiler}
The user profile, shown on the bottom-right of Figure~\ref{fig:personalization_module}, provides a high-level summary of the interactions between the users and the agent to build a dynamic and holistic understanding of the users' preferences. We adopt a hierarchical approach, first proposed in \cite{zhong2023SiliconFriend}, to build the users' profiles. The user profiler starts by grouping all memory entries by date and generating daily user preference summaries to capture user preferences at a high granularity. Next, the daily summaries are aggregated into a single global summary serving as the user profile. Our choice of a hierarchical approach is motivated by two main reasons: (i) scalability: as the long-term memory grows with time, a hierarchical approach is highly scalable because it is amenable to MapReduce-style processing \cite{mapreduce}, (ii) information loss: directly generating a concise summary from the long-term memory involves a long-context prompt. The ability of LLMs to successfully identify relevant information within the input context is known to degrade as the length of the input context increases \cite{liu2023lost}.

\subsubsection{Personalization tool}

The personalization tool (Figure~\ref{fig:personalization_module}) can use the long term memory and the user profiler in order to answer questions about the user's preferences When queried with a question about the user's preferences, this tool first encodes the question into the same embedding space as the memories using the dense retrieval embedding model. Then similar memories are retrieved from the long-term memory using cosine similarity in the embedding space as the distance metric. Next, the tool constructs an LLM prompt consisting of the retrieved memories, the user profile, and the question. Finally, it queries the LLM with the prompt and returns the response. The user profile and the retrieved memories are complementary; the retrieval module identifies a small pool of candidate memories that provide narrow and precise information, while the user profile presents a more holistic view of the user. Multiple users are supported by maintaining separate profiles and long-term memories. User identity is recognized using voice recognition software \cite{vosk}.

\subsubsection{Human interaction tool}
The human interaction tool allows SAGE to ask the user questions, which it usually uses to clarify intent. The tool is called with a string which is communicated to the user via a text-to-speech interface. The tool blocks until the user replies, transcribing their speech to text using an off-the-shelf model such as \cite{radford2022whisper}. Empirically, we have found that the introduction of the user interaction tool can cause the agent to become over-cautious, using this tool over other data sources to reduce uncertainty. We use prompt engineering in SAGE to encourage it not to over-use the tool, but how to best trade off personalization, human interaction, and risk aversion is a topic of active research.

\subsection{Device Interaction}
\label{sec:code_free_smartthings}

In this section, we introduce the set of tools that SAGE uses to interact with smart devices. These tools enable flexible device interaction that is scalable, in the sense that new devices can be integrated into the system with negligible extra development effort and its capabilities can be leveraged to the fullest extent without the need for device-specific code. For example, if a user adds a smart fridge to their smart home ecosystem, SAGE can integrate the presence of the fridge and all of its capabilities (e.g. temperature settings, door opening detection, power consumption, etc.) into its decision making process without the need for any fridge-specific code to be written.

Most smart home systems (SmartThings, Home Assistant, Google Home, Alexa, etc.) provide APIs for interacting with the smart devices in users' homes. These APIs are documented online, and code examples for using them are available, meaning that LLMs trained on web data (such as GPT4) are likely to have some inherent knowledge of these APIs. In practice, we have found that LLMs often fail to use these APIs successfully due to minor errors such as forgetting the exact names of the attributes they need to retrieve. Furthermore, some devices have custom functionality for which no documentation is available online, and can only be retrieved by querying the device's API. These challenges can be overcome if details of API usage are injected into the prompt, but injecting the full documentation for all connected devices in not feasible, as doing so would far far exceeding the maximum prompt length of today's LLMs.

Motivated by LangChain's OpenAPI toolkit \cite{langchain_openapi}, the device interaction tool is implemented as an agent-tool. This agent-tool generates a high-level plan using only a general description of devices and their associated capabilities, retrieves detailed documentation for the subset of capabilities that are required by the plan, then uses these to construct API calls. This behavior is enabled through a collection of tools detailed below.

\textbf{Device interaction planner tool}: Generates a sequence of steps that must be performed by the device interaction agent-tool to complete the given command. The tool is implemented using a single LLM query. This query includes a list of devices, their capabilities, and short descriptions of what each capability does. It also includes the input command and a description of how the plan should be structured. The query specifies that each step of the generated plan should include one or more device IDs, one or more capabilities, and a natural language description of what needs to be done in that step. If the planner cannot directly infer the correct device from the information it has been given, it can supply multiple candidate devices and / or capabilities to the device disambiguation tool (detailed below) to decide on the correct one. The Device Disambiguation tool retrieves detailed documentation for each proposed capability, giving the agent sufficient information to make a final choice.
The use of a planning tool, as opposed to relying on the chain-of-thought planning (described in Section \ref{sec:system_overview}) of the device interaction agent-tool, is motivated by the fact that this planning process requires a large amount of information injected into the prompt. Adding of all of this information directly to the device interaction agent-tool prompt would lead to significantly higher LLM query costs, as the agent-tool prompt is usually called many times within the course of a single use of the device interaction tool.

\textbf{API documentation retrieval tool}: Retrieves documentation about a requested device's capabilities. The documentation is scraped from the web when available, otherwise it is retrieved from the device using the API. While documentation extracted from the device API is often lacking detailed natural language descriptions of usage, it contains the names of attributes, commands, and command arguments, the meaning of many of which can be inferred from the name alone. This tool takes as input a list of capabilities, and returns detailed documentation for each in JSON format. The JSON format is used for convenience, since this is the format returned by the documentation scraper and device APIs. 

\textbf{Device attribute retrieval tool and device command execution tool}: These tools allow the agent to communicate with the API to read attributes and execute commands. We implement these tools as wrappers around the SmartThings REST API to query and modify device states \cite{smartthings_api}. In order to use these tools, the documentation retrieval tool must first be called in order to retrieve the capability details and format the inputs properly. Note that in the event that the inputs are not formatted properly and the API throws an exception, we have found empirically that if the text associated with this exception is propagated back to the device interaction tool agent, it can often react and correct the API request accordingly.

\begin{figure}[t]
    \centering
    \includegraphics[width=0.45\textwidth]{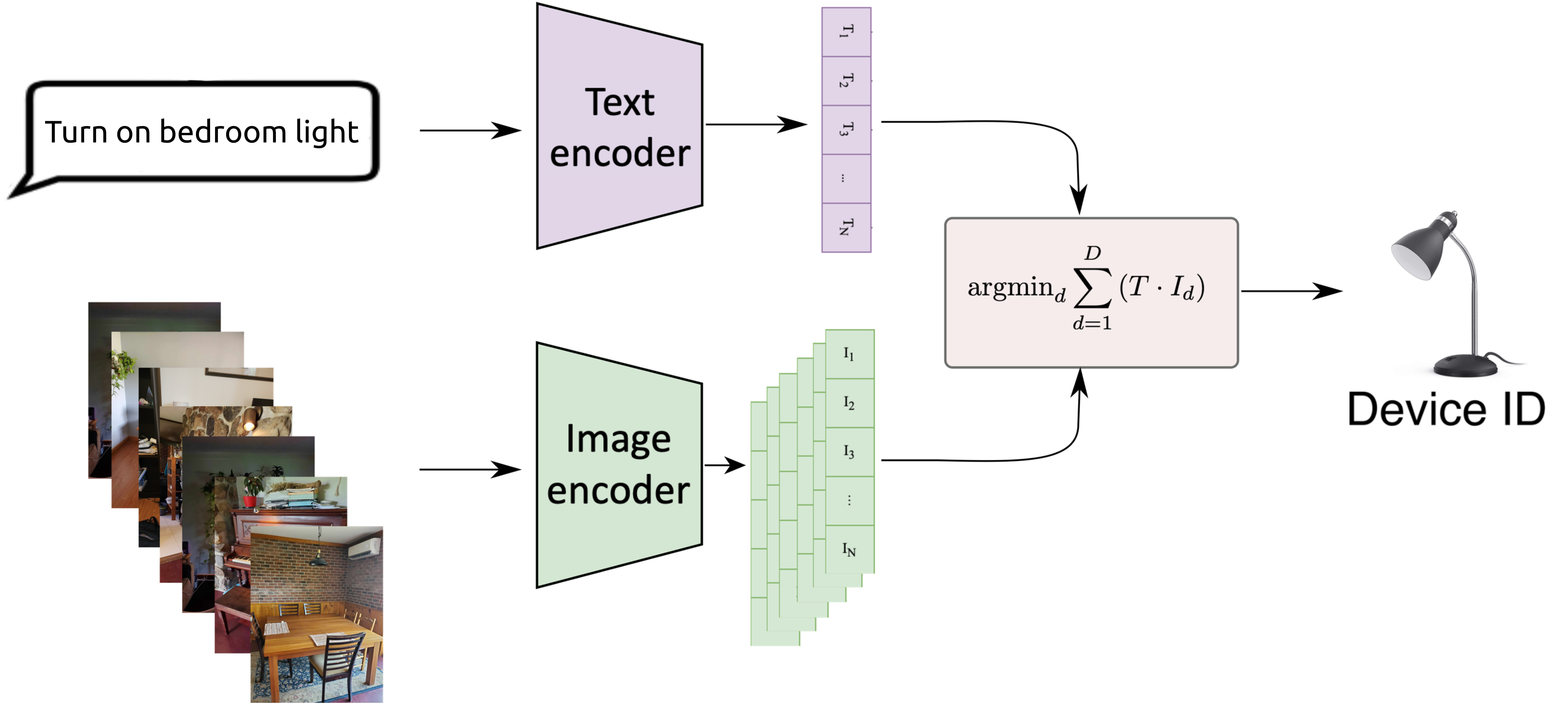}
    \caption{\textbf{Device disambiguation}. A method to resolve the device that best fits a user natural language device description using VLMs.}
    \label{fig:device_disambiguation}
\end{figure}

\textbf{Device disambiguation tool}: Allows the system to resolve which devices the user wants to control in scenarios when there is more than one instance of a given device (e.g. multiple smart lights). We propose a method that can determine  which device is relevant to the task by leveraging visual context. By capturing a photograph of the device within its surroundings (during initial device setup), we can resolve the device ID without requiring the user to hard-code a unique device name (which can easily be forgotten and may not be known to guests). For example, in Figure~\ref{fig:device_disambiguation}, it is obvious from the picture alone that the light is located in the dining room.  The device identity is disambiguated using a Visual Language Model (VLM), as shown in Figure~\ref{fig:device_disambiguation}, where a multimodal VLM (OpenClip ViT-B-32 laion2b\_s34b\_b79k, \cite{open_clip}) is used to compute embeddings for the user's natural language description of the device and each of the device photographs. The device whose image embedding has maximum cosine similarity to the text embedding of the device's description is selected. 

\subsection{Monitoring}
\label{sec:persistant_commands}

This section introduces a set of tools whose functionality relates to device state monitoring. Many of the more powerful smart home behaviors are unlocked by the ability to monitor the state of some device and react to state changes. These behaviors are referred to as persistent commands \cite{king2023sasha} or routines, as the system should \textit{persistently} behave in a desired manner following a conditional event (e.g. the coffee machine should turn on whenever the morning alarm rings).  
Smart home solutions typically approach this problem using conditional statements in applications such  as IFTTT\footnote{https://ifttt.com/} (IF This Then That). Once the condition is defined, condition checking can be performed using low-cost compute resources. A drawback of this approach is the lack of flexibility afforded by the system because IFTTT routines rely on conditional triggers that must be predefined by the manufacturer and manually activated by the user.

Highly flexible persistent command handling could be implemented within the SAGE architecture simply by periodically running SAGE with the persistent command as input. Each time it is run, the agent could check whether the command is satisfied and if it is execute the desired behavior. This approach, which retains all of the capabilities of the agent architecture and is simple to implement, has the downside of requiring the agent to constantly be running, incurring significant computational costs.

\begin{figure}[ht!]
    \centering
    \includegraphics[width=0.45\textwidth]{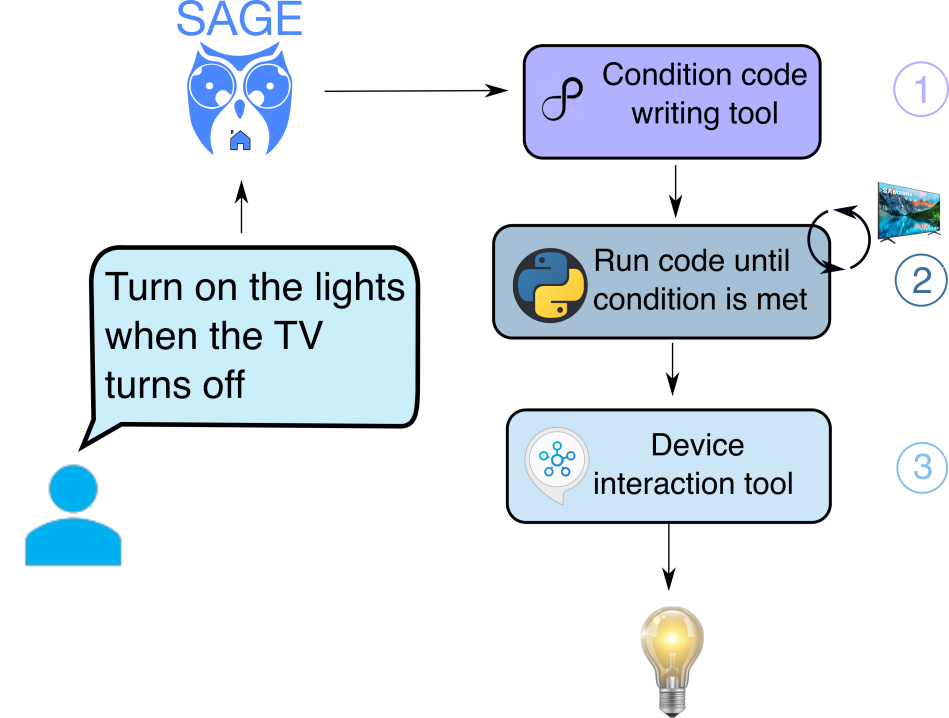}
    \caption{A summary of the persistent command handling mechanism. \textbf{1:} After receiving the user request, the SAGE agent-tool extracts the condition (``is the TV off?") and uses the condition code writing tool to write code to check this condition. The tool registers this code in memory and responds with the name of the function (is\_tv\_off). \textbf{2:} The SAGE agent-tool registers the function is\_tv\_off with the condition polling tool, along with the command reflecting the action to take once the condition is detected ``turn the lights on". At this point the SAGE agent-tool finishes executing. The condition polling tool periodically runs the is\_tv\_off function. Once the function outputs True, the condition polling tool triggers the SAGE agent-tool again with the registered action ``turn the lights on." Note that the agent-tool will only be triggered when the status of the action transitions from False to True to avoid re-running the agent-tool the entire time the TV is off. \textbf{3:} The agent-tool begins executing with the command ``turn the lights on" and turns on the lights using the device interaction tool.}
    \label{fig:persistent_commands}
\end{figure}

In order to increase the system's flexibility while minimizing cost, we propose a method by which SAGE can autonomously program conditional routines by writing python code which implements condition checking logic.  Two tools are introduced to support this functionality: the condition code-writing tool and the condition polling tool. The SAGE agent queries the condition code writing tool to write the necessary code, then registers this code with the condition polling tool, which runs it periodically (once every few seconds). Along with the condition checking code, it also registers a description of the action that must be taken when the condition is met. Once the code returns ``True", the polling process triggers a second execution of the SAGE agent with the command registered with it by the first execution. The entire approach, and an example thereof, is summarized in Figure~\ref{fig:persistent_commands}. 

The implementation of the condition code writing tool is complicated by the same challenge as the device interaction tool -- the requirement to inject API details into the query that generates the code. We overcome this challenge in a similar fashion: by creating an agent-tool which uses the device interaction planner tool and the API documentation retrieval tool. However, instead of the device attribute retrieval and command execution tools (as in the case of the device interaction tool), the condition code writing tool agent-tool has access to a code execution tool which allows it to test its code. This tool also stores the code it has run in memory, so that it can be referred to by the name of the function. Similarly to the device attribute retrieval and command execution tools, the code execution tool handles exceptions by returning their messages to the to the code writing agent-tool, facilitating recovery from faulty code.

\section{Execution example}
\label{sec:execution_example}
Figure~\ref{fig:interactive_demo} visualizes an execution trace of SAGE agent-tool handling a single command, illustrating the sequence of tools that are called to complete the task. After checking long-term user interaction memory and user profile information with the \textit{personalization} tool, the system doesn't find any relevant information and so must call the \textit{human interaction} tool to directly ask the user about their sports preferences. The SAGE agent-tool then finds the relevant TV program and channel using the \textit{TV schedule search} tool. Once the user's concrete goal is established (to change the TV to a particular channel), SAGE calls the \textit{device interaction} tool, an agent-tool. The \textit{device interaction} tool starts by calling the \textit{device interaction planner} tool to establish the sequence of steps that must be performed. The plan it generates is as follows: First, the \textit{device disambiguation} tool is called to determine which TV the user is referring to. Second, the \textit{device command execution} tool turns the TV on. Finally, the \textit{device command execution} tool is used again to change the channel. The \textit{device interaction} tool successfully executes this plan, then returns execution to the top level SAGE agent-tool, which responds to the user and terminates the run.

\section{Evaluation}
\label{sec:evaluation}
To evaluate SAGE and competing methods, we created a dataset of 50 tasks. Test tasks are implemented by initializing device states and memories, running the SAGE agent-tool with an input command, then evaluating whether the device state was modified appropriately. For tasks that involve answering a user's questions (as opposed to modifying device states), the tests are designed such that to answer the question the agent must retrieve a specific piece of information (which it is unlikely to be able to guess). An LLM-based evaluator is then used to check whether the answer contained the expected information. The results of all tests are binary (pass / fail).

The smart device configuration (device types, IDs, etc) was created by configuring a home with real SmartThings devices, then saving the state of these devices. The devices included: 2 televisions, 1 refrigerator, 1 dishwasher, and 4 lights. The initial states of the devices are modified by the initialization routine of each task. Photographs of the real devices in their real locations are used in the device disambiguation tool.

We classify the test cases according to five types of challenges that are difficult for existing systems to handle. Most tests in the set belong to one or more of these categories. The categories are:
\begin{enumerate}
    \item \textbf{Personalization}: Integrating knowledge of user preference to interpret the request correctly.
    \item \textbf{Intent resolution}: Understanding vague commands and drawing logical conclusions.
    \item \textbf{Device resolution}: Identifying the desired device ID based on natural language description.
    \item \textbf{Persistence}: Handling commands that require persistent monitoring of system states.
    \item \textbf{Command chaining}: Parsing a complex command that consists of multiple instructions, breaking it into actionable steps and executing each step in a coherent manner. 
\end{enumerate}

We also include a sixth category, \textbf{Direct command}, to indicate tests cases that are simpler to execute in that they do not feature any of the 5 challenges listed above.

The test cases are designed to be run in a completely automated fashion. For this reason, we disable the human interaction tool during testing. We test SAGE with 5 different LLMs: GPT4, GPT4-turbo, GPT3.5-turbo (ChatGPT) \footnote{https://platform.openai.com/docs/models}, Lemur \cite{xu2023lemur}, and Claude2.1\footnote{https://www.anthropic.com/index/claude-2-1}. For all LLMs, we set the temperature parameter to 0. For each LLM, we run each test case 3 separate times, since LLM performance is somewhat stochastic, even with 0 temperature.

In addition to the main set of 50 tasks, we also created a set of 10 extra ``test set" tasks after the development of SAGE was complete. The aim of these tasks was to verify that the prompts had not been over-engineered for the task set. The author who developed these tasks was familiar with the SAGE architecture, but was not involved in the final prompt engineering stages.

To provide more insights on the key challenges faced by each LLM, we manually annotated the failures that consistently occurred across all three runs. The failures were categorized based on the nomenclature defined in Table \ref{table:failure_category_descriptions}. Since the success of a given test case is contingent upon multiple steps of decision-making, failures can occur in more than one step in the LLM's line of reasoning. For our analysis, the classification of a failure of a given test case is based on the \emph{first} mistake in the execution. 

We contextualize the failure categories by grouping them into tiers. These tiers are organized such that, most of the time, a failure in tier $n$ implies success in tiers $1$ through $n-1$. For example, if an annotator marked a test case as failing due to a failure in planning (tier 2 failure), this implies that it succeeded at command understanding and formatting (tier 1 failures).

\begin{table*}[ht]
\centering
\caption{Description of the failure categories. The failure tiers column helps contextualize the failure type category. In most cases, a failure on tier n implies that failures in lower tiers were avoided (4 = highest, 1 = lowest).}
\label{table:failure_category_descriptions}
\begin{tabular}{|c|l|l|}
\hline
\textbf{Failure Tier} & \textbf{Failure Type} & \textbf{Description} \\ \hline
\hline 
 \multirow{2}{*}{1} & Formatting & Agent fails to follow the template required by ToolInstructions. \\
& Command understanding & Agent fails to understand what the user is asking. \\
\hline 
 \multirow{1}{*}{2} & Planning & Agent's plan is incorrect or missing steps. \\
\hline 
 \multirow{5}{*}{3} & Plan execution & Agent proposes a correct plan but the execution is missing steps.\\
& Tool selection & Agent chooses the wrong tool at a given stage of the execution. \\
& Tool population & Agent provides incorrect arguments to the tool. \\
& API request & Agent makes incorrect SmartThings API request (wrong attribute, command, or component).  \\
& Code writing & Agent's generated code does not work properly. \\
\hline 
 \multirow{3}{*}{4} & Faulty tool & Non-agent failure (usually a failure of the VLM in the device disambiguation tool). \\
& LLM limitation & LLM lacking key common knowledge or context length. \\
& Hallucination & Agent invents concepts (e.g. devices, components, user requirements) that do not exist. \\
\hline 
 \multirow{1}{*}{NA} & Other & All other failures. \\
\hline
\end{tabular}
\end{table*}
 
\subsection{Baselines}
To contextualize SAGE's performance, we compare our method to two LLM-based smart home automation baselines on our test tasks. The first method, called ``One Prompt", involves creating a single prompt comprised of the user command as well as the states of all devices, and asking the LLM to generate updated states in response. The full device state, serialized to JSON format, exceeds GPT4's token limit ($8000$ tokens), so we manually selected the parts of the device state involved in the tests. In addition, the model was asked to output the changes that need to be made, not the full new state.

The second baseline, called ``Sasha," implements the pipeline described in \cite{king2023sasha}, with some modifications. The original pipeline in \cite{king2023sasha} consists of 5 pipeline states -- clarifying, filtering, planning, feedback, and execution. The clarifying and feedback stages required human intervention, and were thus not compatible with our fully automated testing framework, so they were removed in our implementation. Additionally, this pipeline distinguishes between ``sensors" and actionable devices, allowing the pipeline to output sensor-based trigger-action pairs to handle persistent commands. This requires the manual definition of triggers, which our testing framework does not support, since SAGE is able to generate its own triggers by writing code. As such, our implementation of Sasha does not include the trigger concept, and is therefore unable to handle persistent commands.

Both of these baselines are at a disadvantage in that they are not able to integrate all of the different sources of information that SAGE uses (e.g. user preferences, photos of the devices, etc.). Despite this, the baselines allow the reader to gauge the difficulty of the task set, and to appreciate the extent to which integrating information from a variety of sources can improve the performance of smart home automation systems. We do not provide a baseline with access to the same information as SAGE because, to our knowledge, there is no previous work that is capable of integrating all of these information sources.

\section{Results}
\label{sec:results}
\begin{figure}[t]
    \centering
    \includegraphics[width=0.49\textwidth]{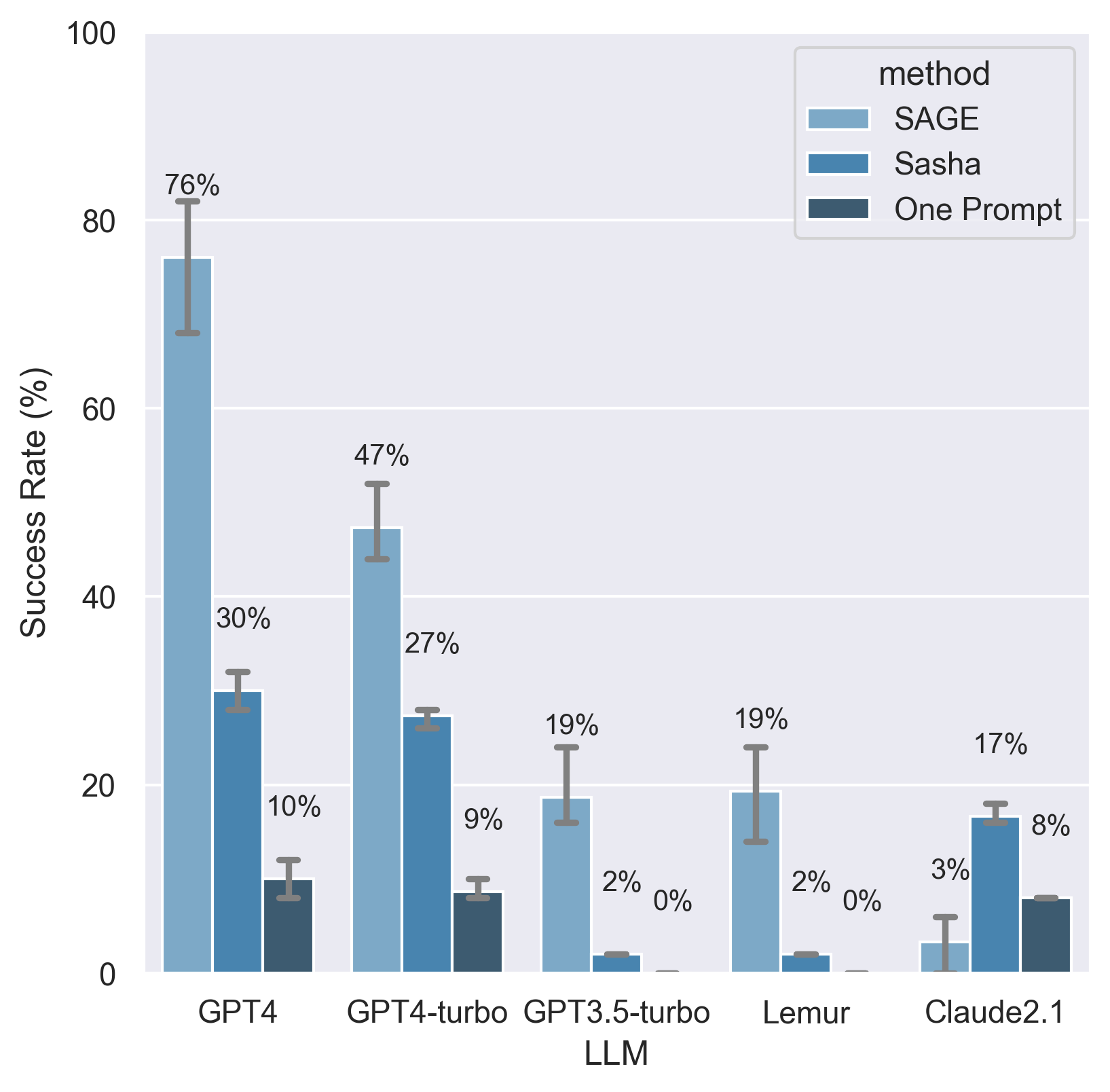}
    \caption{Overall success rates on 50 challenging tasks. SAGE leverages a collection of tools that allow it to integrate a large amount of information into its decision making process, allowing it to outperform other LLM-based methods by a significant margin. Designed primarily for use with GPT4, it outperforms other methods with a variety of LLMs, including open source models such as Lemur. Error bars indicate max and min scores over the three runs.}
    \label{fig:sage_results}
\end{figure}

\begin{figure}[t]
    \centering
    \includegraphics[width=0.49\textwidth]{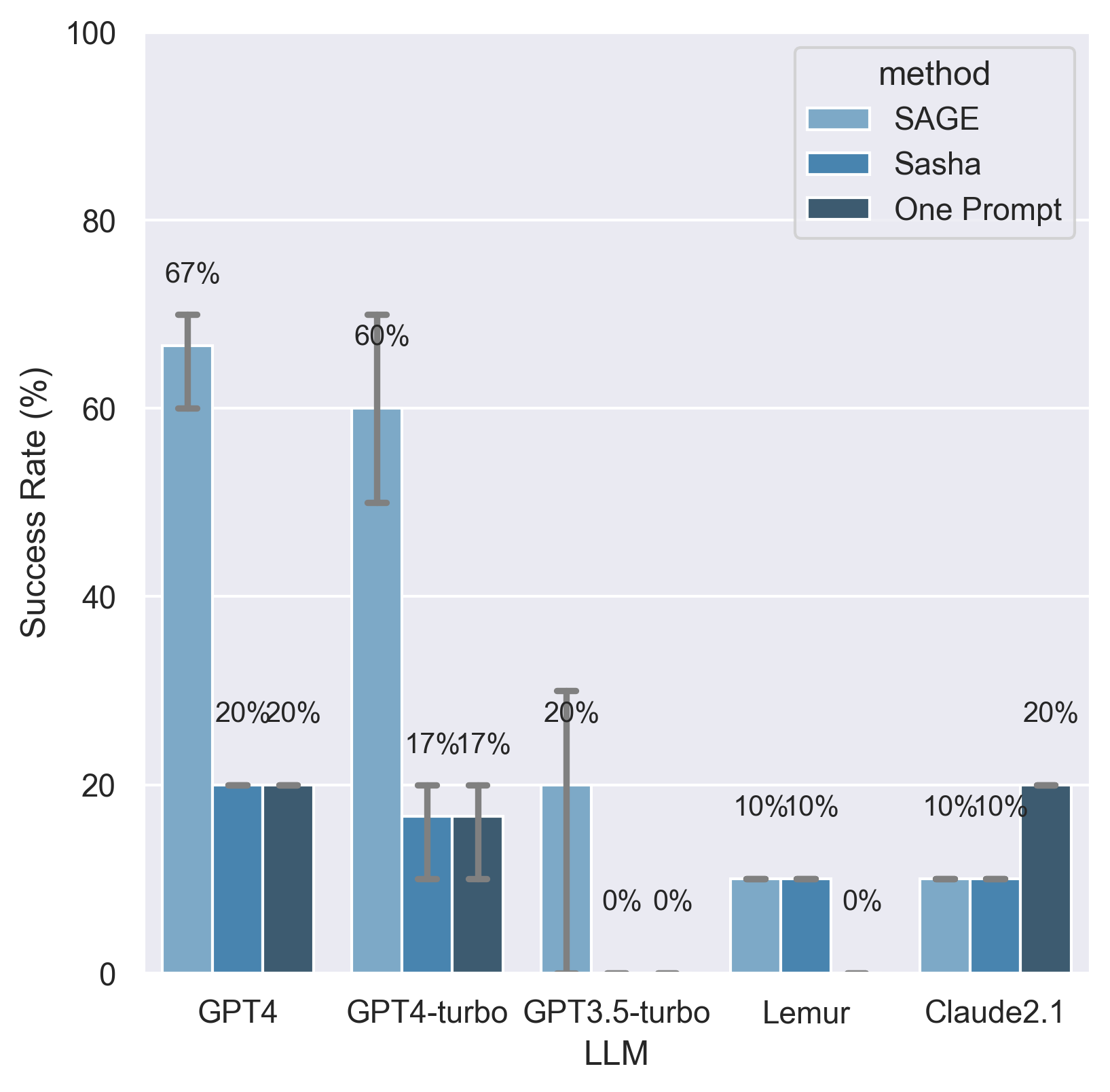}
    \caption{Overall success rates on a ``test set" of 10 extra tasks. These tasks were not seen by the SAGE's designers during the iteration process, and serve to validate that the prompts used in SAGE are not over-engineered to a particular task set. Error bars indicate max and min scores over the three runs.}
    \label{fig:results_test_set}
\end{figure}

Overall success rates for the three methods, SAGE, Sasha, and One Prompt on the 50 task set are presented in Figure~\ref{fig:sage_results}. SAGE achieves an overall success rate of 75\% with GPT4, far beyond either of the baselines, demonstrating that it is indeed capable of integrating a variety of information sources through the use of its tools. Unsurprisingly, GPT4 achieves the highest performance of all LLMs regardless of method. This is largely because it is the most powerful, but also because the prompts were optimized for it. GPT4-turbo appears to be significantly worse than GPT4 on the 50 task set, though it is purported to have similar performance. There are several potential explanations for this phenomenon. First, the details of this model have not been publicly released, but based on the fact that it is several times less expensive to run, we can guess that it has fewer parameters than GPT4. Second, our experience has been that GPT4-turbo is more conversational than GPT4, making it less likely to follow instructions, and therefore worse at agent-style tasks. Finally, the SAGE prompts are optimized for GPT4. It is likely that if we invested equal effort in optimizing them for GPT4-turbo, its performance would increase significantly. On the 10-task ``test set" illustrated in Figure~\ref{fig:results_test_set}, the two models attain approximately equal performance. However, given the small size of this set, it may be less accurate in resolving performance differences than the primary set.

Another interesting observation from Figure~\ref{fig:sage_results} is that Claude2.1 performs very poorly with SAGE, even compared to both baselines. This can primarily be attributed to Claude2.1's inability to follow formatting instructions and poor tool selection abilities, as illustrated by Table~\ref{table:manual_analysis}, which presents the results of the manual failure analysis. In order to successfully function as the backbone of an agent, an LLM must be capable of following instructions related to formatting and tool selection, as an agent must be capable of converting natural language reasoning into concrete real-world actions. Claude2.1 does not seem to have significant reasoning issues (as evidenced by a lack of failure in the planning stages), but its lack of ability to perform this basic functionality makes it a poor candidate for SAGE and other agent architectures.

Parity of GPT3.5-turbo and Lemur is an encouraging result for those interested in the use of open source LLMs for agent applications.

\begin{figure}[t]
    \centering
    \includegraphics[width=0.49\textwidth]{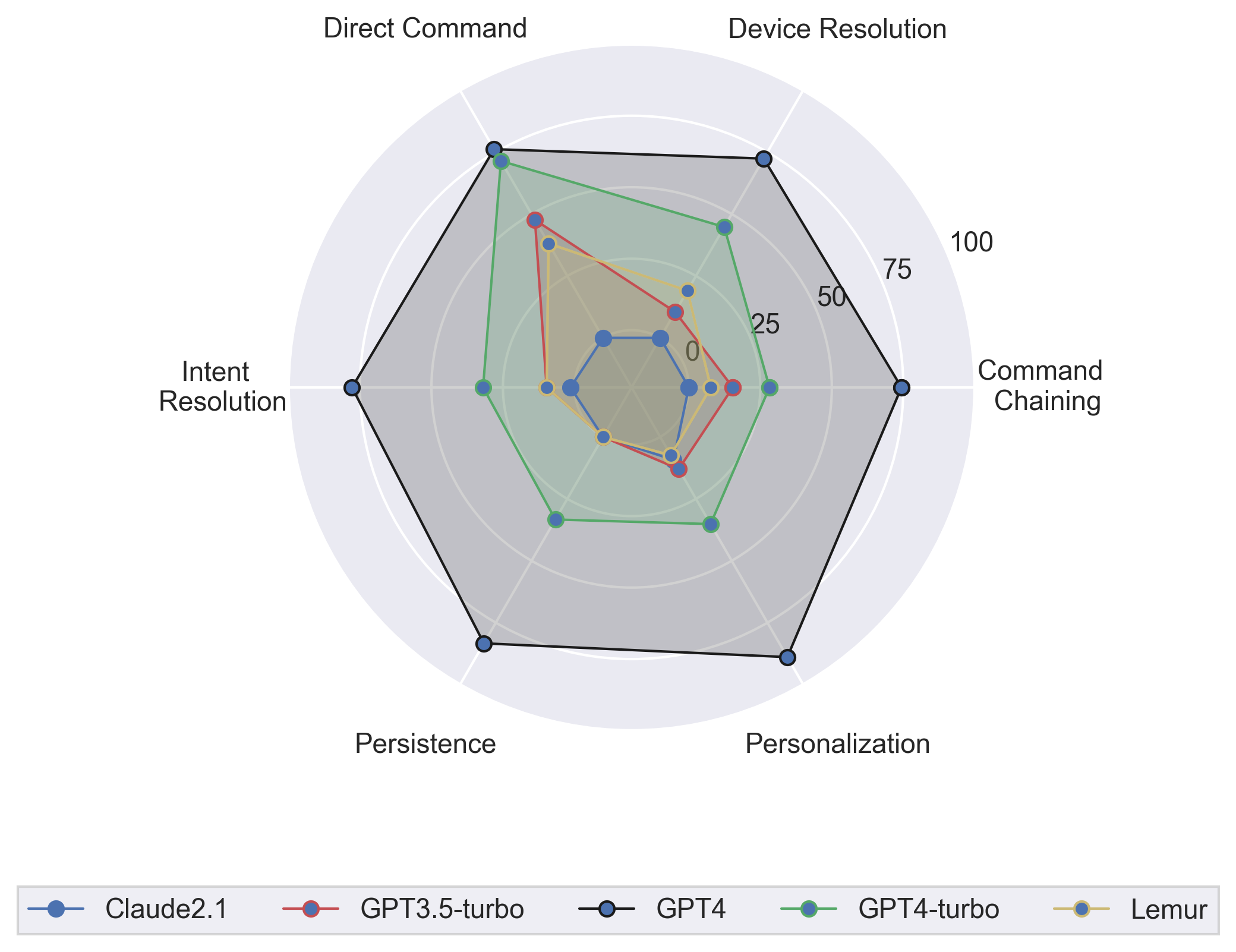}
    \caption{SAGE success rates for each category of challenging tasks on the 50 task set. Note the inner ring of the chart indicates zero success rate, not the center.}
    \label{fig:results_categorical}
\end{figure}

Examination of Figure~\ref{fig:results_categorical}, which presents SAGE success rates for each category of challenging tasks, reveals that SAGE with GPT4-turbo maintains fairly consistent performance across challenge categories. On the other hand, lesser LLMs tend to perform much better on direct commands than more challenging tasks. This confirms that the categories we have identified do indeed present significant difficulties for smart home automation systems.

\subsection{Failure analysis}

\begin{table*}[ht]
\centering
\caption{Results of manual failure analysis. This analysis categorizes failures into one of 12 failure types. The failure tiers column helps to contextualize the failure type category. In most cases, a failure in tier n implies that failures in lower tiers were avoided (4 = highest, 1 = lowest). Only test cases that failed consistently across all three runs were analyzed.}
\begin{tabular}{|c|l|c|c|c|c|c|}
\hline

\textbf{Failure Tier} & \textbf{Failure Type} & \textbf{GPT-4} & \textbf{GPT-4 Turbo} & \textbf{GPT-3.5 Turbo} & \textbf{Lemur} & \textbf{Claude 2.1} \\ 
\hline
& &  \multicolumn{5}{c|}{Total Failures = \# failures per run $\times 3$}\\
\cline{3-7}
& &  9 & 51 & 96 & 111 & 141 \\
\cline{3-7}

& &  \multicolumn{5}{c|}{Failure rate (\%)}\\
\hline 
 \multirow{2}{*}{1} & Formatting & 0 & 0 & 12 & 0 & 35  \\
& Command understanding & 67 & 14 & 2 & 0 & 1  \\
\hline 
 \multirow{1}{*}{2} & Planning & 0 & 6 & 32 & 43 & 0  \\
\hline 
 \multirow{5}{*}{3} & Plan execution & 0 & 2 & 5 & 10 & 2  \\
& Tool selection & 0 & 4 & 8 & 3 & 37  \\
& Tool population & 0 & 18 & 27 & 24 & 15  \\
& API request formatting & 33 & 51 & 0 & 0 & 1  \\
& Code writing & 0 & 6 & 6 & 0 & 0  \\
\hline 
 \multirow{3}{*}{4} & Faulty tool & 0 & 0 & 5 & 12 & 0  \\
& LLM limitation & 0 & 0 & 1 & 7 & 0  \\
& Hallucination & 0 & 0 & 0 & 0 & 2  \\
\hline 
 \multirow{1}{*}{NA} & Other & 0 & 0 & 0 & 1 & 7  \\
\hline
\end{tabular}
\label{table:manual_analysis}
\end{table*}
 
Table~\ref{table:manual_analysis} summarizes the results of the manual failure analysis. In this section we review each of the LLMs and provide some discussion about the primary reasons they fail.

Out of GPT4's $9$ failures that were analyzed, $6$ are due to misunderstandings of the user command. These commands are somewhat vague, and the LLM was not able to apply the common sense expected by the test design. The other $3$ failures were due to the LLM using the wrong device component (in the SmartThings API some devices have multiple ``components", e.g. one for the freezer part and one for the refrigerator part of a smart fridge).

GPT4-turbo performance is hindered by tool usage issues, which account for $69\%$ of the failures (API request formatting and tool population). Indeed, GPT4-turbo most often fails due to problems formulating a request that respects the SmartThings API. This includes providing the right argument types, components, and capabilities.

GPT3.5-turbo predominantly fails to generate proper plans and struggles with passing the right parameters to tools. It also usually fails to use the state of the available devices to inform its decision making process. It also often fails to correctly interpret the user's request, which tends to require significant common-sense reasoning abilities. 

Lemur struggles to correctly interpret the user command, given the context of current device state, and thus also struggles to generate good plans. The plans generated by Lemur are missing vital steps like using device disambiguation to find the right tool or recognizing the need to create a trigger for a persistent command. It also utilizes multiple rounds of trial and error to find the right API calls for device control.

Claude2.1 often struggles to follow the expected format of the response. This involves outputting a thought followed by an action based on previous observations. Although we provide verbose error log and debugging hints to the agent, Claude2.1 fails to recover from its mistakes. Another common failure type for Claude is tool selection. For example, Claude2.1 often uses the personalization tool for direct commands - where it is not needed - instead of the device interaction tool. Note that tool use is a new feature recently added to Claude and still under development.

\section{Conclusion}
This article introduced SAGE, an LLM agent architecture targeted at smart home applications. SAGE orchestrates the use of tools in a sequential decision making process. SAGE integrates a collection of novel tools designed to tackle key challenges in smart home automation, including personalization, flexible device interaction, device monitoring, and device disambiguation. We also created a dataset of challenging smart home automation test cases which tested the system's ability to be personalized, to resolve user intent from unstructured queries, to resolve devices referred to in natural ways (e.g.``the light by the piano"), and to appropriately handle command persistence and chaining. These challenges are very difficult for today's smart home automation systems, but reflect the sophistication smart home users will demand next-generation systems. 

SAGE achieved a success rate of around $75$\% on these tasks. This value, while imperfect, is much higher than that achieved by existing home automation solutions, even those based on powerful LLM technology. Each success required the successful sequential use of many tools, meaning that in fact the number of successful tool uses is much larger than the number of failed ones. To understand the underlying causes of failures, we manually analyzed and categorized each one. This analysis revealed that there are many ways that such a complex decision-making process can go wrong, and that today's most powerful LLMs are needed in order to achieve acceptable performance. However, the respectable performance attained by Lemur, an open-source model fine-tuned for agent tasks, is encouraging. We expect that in one or two years the performance of SAGE with an open source LLM will be comparable to that of GPT4 today, at a fraction of the cost.

As such, SAGE represents a promising first step towards the creation of truly flexible smart home automation systems that users can interact with as naturally as they would with a close friend.

\section{Related Work}
\subsection{Home Automation}
A smart home system consists of connected IoT (Internet of Things) smart devices that enable the simultaneous monitoring, sensing, and control of the home environment. Automating the the control of these devices can lead to improvements in quality of life, comfort, and resource usage \cite{ki2020can}. Recent work has aimed to used machine learning to enhance the capabilities of smart home systems. For example, \cite{manu2019smart} proposed to perform home automation based on activity recognition, developing a deep learning algorithm to recognise users' activities from accelerometer data. Another focus is on voice-based home assistants, where the system is tasked with understanding users' voice utterances. For instance, \cite{8261311} developed a voice controlled home automation system based on NLP (Natural Language Processing) and IoT to control four basic appliances. 
Leveraging advanced NLP techniques, current commercial solutions such as Bixby, Google Assistant, and Alexa offer a user-friendly interface capable of handling a variety of commands and questions from shopping and setting reminders to device control and home automation. However, these modern home assistants usually struggle with implicit and complex commands \cite{luger2016like}. In other words, these systems tend to fail in situations where a user utterance can not easily be mapped to a pre-programmed routine.

In an attempt to overcome some of these challenges, recent work proposed to leverage the reasoning capabilities of LLMs to better understand and carry out user commands. In particular, Sasha \cite{king2023sasha} introduced the use of LLMs in smart home environments and showcased that the LLMs can be used to produce reasonable behaviors in response to complex or vague commands. Sasha implements a decision making pipeline where each step (such as selecting the device to use, or checking if a routine already exists) is implemented using an LLM. However, unlike SAGE, the stages of the Sasha pipeline are manually defined and fixed, limiting its flexibility. In contrast, SAGE relies on an LLM to decide the sequence of steps to perform.

Today's smart home users also have the option of manually defining IFTTT-style routines, which connect trigger conditions to actions \cite{DBLP:journals/corr/abs-2110-00068}, in order to create complex behaviors. This approach is inconvenient because it must be implemented by users manually through an app or web interface. It is also limited by the fact that trigger conditions must be defined by device manufacturers. Users can also write their own apps to manage device states, such as with SmartThings SmartApps \cite{smartapps}, but this requires a level of technical sophistication beyond the ability of most users, as well as a significant time investment.
Web-based services such as IFTTT, Zapier\footnote{https://zapier.com/}, and Home Assistant\footnote{https://www.home-assistant.io/} enable the user to create rules to control their smart devices. The advantage of these services is that they simplify the process of connecting various services and smart devices without the need for extensive programming knowledge. However, these solutions also lack the reasoning and context-awareness offered by LLMs.

Recently, IFTTT has begun to leverage LLMs through the creation of a ChatGPT plugin\footnote{https://ifttt.com/explore/business/ifttt-ai}, which provides a more user-friendly and accessible interface to interact with the automation platform. However, this plugin does not allow ChatGPT to generate new routines, rather only to trigger existing ones. Zapier also offers an LLM-based offering, which allows users to create automations in a more natural way, but currently lacks significant reasoning sophistication. \cite{king2023sasha} created an LLM-based pipeline which outputs trigger-action pairs to create simple IFTTT routines, but the logical complexity of these routines, as well as the flexibility of the triggers, is limited. Earlier academic work investigated training sequence-to-sequence models to synthesize IFTTT or Zapier routines from natural language descriptions \cite{DBLP:journals/corr/abs-2002-03485}. Results were promising but limited in that the sequence models were able to generate the sequence of functions to call, but not the arguments to those functions.

\subsection{Autonomous Agents}

LLM-powered autonomous agents are designed to to perform complex and diverse tasks. Usually, this involves decomposing the task into multiple stages or subtasks. Several agent architecture designs have been proposed in the literature \cite{LLAsSurvey}. Chain-of-Thought (CoT) \cite{wei2022cot} is a well-known prompting technique that enables the agent to perform complex reasoning through step-by-step planing and acting. In the CoT implementations, several CoT demonstrations are inserted in the prompt to guide the agent's reasoning process. Alternatively, zero-shot CoT \cite{kojima2022zero-shot-cot} demonstrated the reasoning capabilities of LLMs by simply adding the sentence ``think step by step" in the prompt. Another line of work extended CoT by adopting a tree-like reasoning structure where each intermediate step can have its own set of sub-steps (e.g. \cite{yao2023tot}). The aforementioned work did not consider environment feedback (i.e. the outcomes of actions already taken) in the plan generation process. Subsequent work did incorporate this feedback. For example, ReAct agent \cite{yao2022react} incorporates observations from the environment (e.g., outcomes of API calls or tools) received after taking an action. These observations are taken into consideration in each subsequent reasoning step. Human feedback can also help the agent adapt and refine its plan by asking for more details, preferences etc.
 
Another important part of the agent design is the use of external tools for action execution. These enable the agent to go beyond its internal knowledge. APIs are the most common type of tool, and LLMs (such as  Gorilla \cite{patil2023gorilla} and ToolLLM \cite{qin2023toolllm}) have been trained specifically for API use. In addition to APIs, external knowledge bases (e.g. databases of documents) can be used as a tool to acquire specific information or expert knowledge \cite{ge2023openagi}.

{
\bibliographystyle{IEEEtran}
\bibliography{references.bib}
}

\vfill
\end{document}